\pdfoutput=1

\documentclass[11pt]{article}

\usepackage{acl}

\usepackage{times}
\usepackage{latexsym}

\usepackage[T1]{fontenc}

\usepackage[utf8]{inputenc}

\usepackage{microtype}

\usepackage{hyperref}       
\usepackage{booktabs}       
\usepackage{amsfonts}       
\usepackage{nicefrac}       
\usepackage{pifont}
\usepackage{tabularx}
\usepackage{multirow}
\usepackage{graphicx}
\usepackage{amssymb}
\usepackage{amsmath}
\usepackage{eucal}
\usepackage{bm}
\usepackage{subfigure}
\usepackage{enumerate}
\usepackage{color}
\usepackage{caption}
\usepackage{titlesec}
\usepackage{enumitem}
\usepackage{xcolor}
\usepackage{cellspace}
\usepackage{wrapfig}
\usepackage{lipsum}
\usepackage{float}

\usepackage[linesnumbered, ruled]{algorithm2e}
\usepackage{algpseudocode}

\SetArgSty{textup}

\usepackage{hyperref}
\makeatletter
\def\UrlAlphabet{%
      \do\a\do\b\do\c\do\d\do\e\do\f\do\g\do\h\do\i\do\j%
      \do\k\do\l\do\m\do\n\do\o\do\p\do\q\do\r\do\s\do\t%
      \do\u\do\v\do\w\do\x\do\y\do\z\do\A\do\B\do\C\do\D%
      \do\E\do\F\do\G\do\H\do\I\do\J\do\K\do\L\do\M\do\N%
      \do\O\do\P\do\Q\do\R\do\S\do\T\do\U\do\V\do\W\do\X%
      \do\Y\do\Z}
\def\UrlDigits{\do\1\do\2\do\3\do\4\do\5\do\6\do\7\do\8\do\9\do\0}
\g@addto@macro{\UrlBreaks}{\UrlOrds}
\g@addto@macro{\UrlBreaks}{\UrlAlphabet}
\g@addto@macro{\UrlBreaks}{\UrlDigits}
\makeatother

\definecolor{red0}{RGB}{241,65,36}
\definecolor{blue0}{RGB}{16,156,206}
\definecolor{purple0}{RGB}{42,60,196}

\newcommand\modelname{{FacTree}}

\title{Fact-Tree Reasoning for N-ary Question Answering \\
over Knowledge Graphs}

\author{
Yao Zhang$^1$ \ Peiyao Li$^1$ \ Hongru Liang$^2$ \ Adam Jatowt$^3$ \ \textbf{Zhenglu Yang}$^{1}$\thanks{~~Corresponding author.}\\
  \normalsize{$^1$TKLNDST, CS, Nankai University, China, $^2$Sichuan University, China, } \\
  \normalsize{$^3$University of Innsbruck, Austria } \\
  \normalsize{\texttt{$\left \{ \right .$yaozhang, peiyao\_li$\left.\right \}$@mail.nankai.edu.cn, lianghongru@scu.edu.cn, }}\\ 
  \normalsize{\texttt{adam.jatowt@uibk.ac.at, yangzl@nankai.edu.cn}} \\}

\begin{document}
\maketitle
\begin{abstract}

  Current Question Answering over Knowledge Graphs~(KGQA) task mainly focuses on performing answer reasoning upon KGs with binary facts. However, it neglects the $n$-ary facts, which contain more than two entities.
In this work, we highlight a more challenging but under-explored task: $n$-ary KGQA, i.e., answering $n$-ary facts questions upon $n$-ary KGs.
Nevertheless, the multi-hop reasoning framework popular in binary KGQA task is not directly applicable on $n$-ary KGQA.
We propose two feasible improvements: 1) upgrade the basic reasoning unit from entity or relation to fact, and 2) upgrade the reasoning structure from chain to tree.
Therefore, we propose a novel fact-tree reasoning framework, \modelname, which integrates the above two upgrades.
\modelname~transforms the question into a fact tree and performs iterative fact reasoning on the fact tree to infer the correct answer.
Experimental results on the $n$-ary KGQA dataset we constructed and two binary KGQA benchmarks demonstrate the effectiveness of \modelname~compared with state-of-the-art methods.


\end{abstract}

\section{Introduction}
\label{sec:intro}

The task of Question Answering over Knowledge Graphs~(KGQA) has provided new avenues to the
recent development of QA systems 
by utilizing the advantages of KGs~\cite{yu-etal-2017-improved,dubey2019lc,huangwsdm2019, zhang2020neural}. 
Current KGQA studies mainly consider performing answer reasoning upon KGs with binary facts, which encode binary relations between pairs of entities, e.g., \textit{\underline{Golden State Warriors}' arena is \underline{Chase Center}}\footnote{Entities are underlined.}.
However, $n$-ary facts that involve more than two entities are also ubiquitous in reality~\cite{NaLP,Abboud2020BoxE,Wang2021link}, e.g., the ternary fact \textit{\underline{Golden State Warriors} won \underline{the NBA championship} in \underline{2018}}.
Compared to binary facts, $n$-ary facts have more information content.
This makes the answer reasoning for questions involving $n$-ary facts more intractable, exposing open challenges in KGQA.
In this work, we aim to study the under-explored \textbf{$\bm n$-ary KGQA} task, i.e., answering $n$-ary facts questions upon $n$-ary KGs.

\begin{figure*}[t]
    \centering
    \includegraphics [width=0.99\textwidth]
    {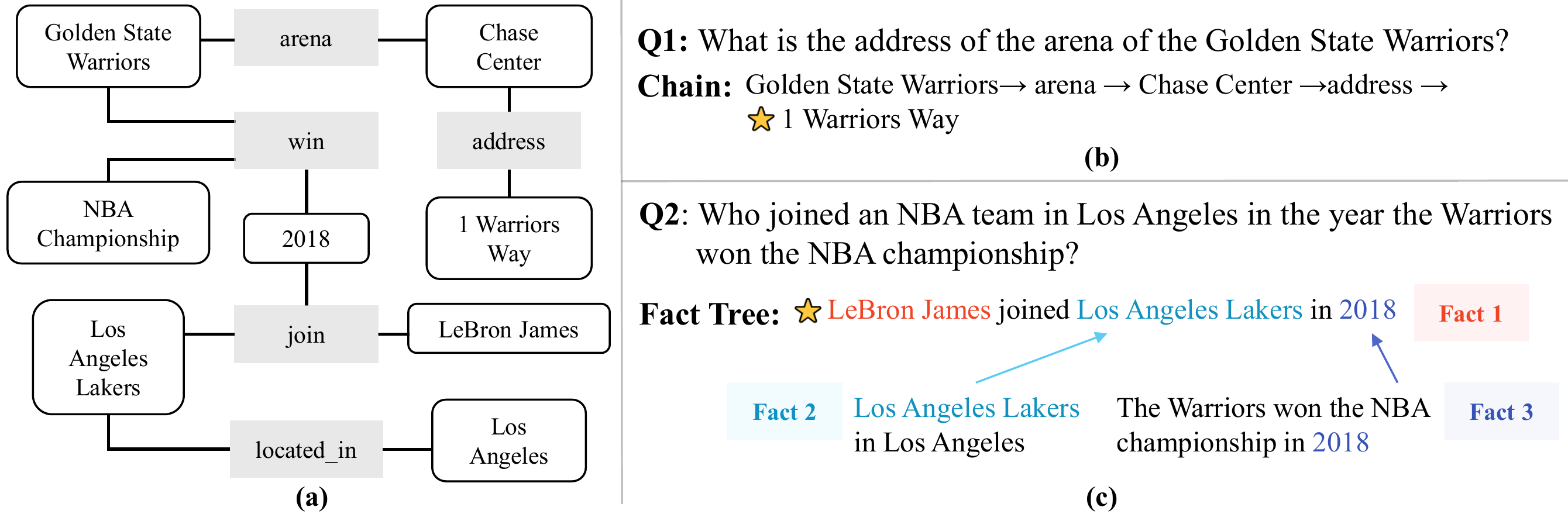}
    \caption{
        (a) A KG fragment, where entities are represented by round rectangles. \texttt{Win} and \texttt{join} are two ternary relations.
        (b) and (c) are two QA examples, where the correct answer is marked by a star.
        The multi-hop reasoning method can be used to answer Q1, and the reasoning process can be visualized as a chain~(b).
        However, for the more complex Q2, the multi-hop reasoning method is not applicable.
        We use fact as the basic reasoning unit to construct the reasoning process.
        As shown in (c), the reasoning process can be visualized as a fact tree: {\color{blue0}\bf fact 2} and {\color{purple0}\bf fact 3} are leaf nodes need to be inferred first, and then the two inferred entities~(\texttt{Los Angeles Lakers} and \texttt{2018}) are transmitted to the root node ({\color{red0}\bf fact 1}). Finally the root node infers the answer entity~(\texttt{LeBron James}).
    }
    \label{fig:intro}      
    \end{figure*}

The multi-hop reasoning KGQA method~\cite{das2017go,Qiu2020Stepwise,saxena2020improving,pmlrv139ren21a} has become popular for its high efficiency and interpretability. 
Specifically, the reasoning process can be expressed as a chain, starting from an entity extracted from the question and then walking on the KG by connected relations and entities until arriving at the answer entity.
See Figure~\ref{fig:intro} (b) for an example, to answer the question, \textit{what is the address of the arena of the Golden State Warriors}, the reasoning chain starts from \texttt{Golden State Warriors}, to walk through ``arena$\rightarrow$Chase Center$\rightarrow${address}", and it ends at \texttt{{{1 Warriors Way}}}, i.e., the answer.
Multi-hop reasoning has been studied widely on the binary KGQA task.
Here, we first try to execute it on the $n$-ary KGQA task.

However, we find that multi-hop reasoning is not directly applicable on $n$-ary KGQA.
We take the $n$-ary facts question in Figure~\ref{fig:intro} (c) as an example to explain.
First, the essence of multi-hop reasoning is to construct a reasoning chain by treating the relation as the translation between two entities~\cite{bordes2013translating,DBLP:conf/nips/RenL20}, naturally in a linear structure.
However, 
the transition from binary to $n$-ary facts is similar to the transition from a line to a plane.
For a single $n$-ary fact (e.g., \textit{Golden State Warriors won the NBA championship in 2018}), the reasoning chain could only include two entities~\texttt{Golden State Warriors} and \texttt{NBA championship} and a relation~\texttt{win} involving them, leading to the possible loss of important information~\texttt{2018}.
To overcome this weakness,
we propose that \textbf{upgrading the basic reasoning unit from an entity or relation to the fact} to expand the coverage of information during reasoning.

Nevertheless, multi-hop reasoning would still be less capable in more complex reasoning scenarios where a question involves multiple $n$-ary facts.
For example, the question in Figure~\ref{fig:intro} (c) is composed of three facts.
When using fact as the basic reasoning unit, the whole reasoning process can be represented as a tree structure, where nodes represent facts and edges reflect the reasoning order. 
Specifically, in the fact tree, the entities~(\texttt{Los Angeles Lakers} and \texttt{2018}) which are missing in the two leaf nodes~(fact 2 and fact~3) are first inferred and then passed to the root node~(fact 1).
The root node can then finally infer the correct answer entity~\texttt{LeBron James}.
Obviously, the chain structure used in the multi-hop reasoning framework is evidently insufficient to cope with the tree structure.
Therefore, to improve the ability to cope with more complex reasoning scenarios, we propose that \textbf{upgrading the reasoning structure from chian to tree}.


In this work, we propose a novel \textbf{fact-tree reasoning framework}, namely, \textbf{\modelname}, which integrates the above two upgrades and pipelines the answer reasoning process into three steps:
1) fact tree construction, which transforms an input natural language~(NL) question into an NL fact tree; 
2) fact location, which locates the NL fact onto the KG;
and 3) fact reasoning, which iterates intra-fact and inter-fact reasoning to infer the answer.
During the intra-fact reasoning, the $n$-ary KG embedding model~\cite{guan2020neuinfer} is plugged in to alleviate the deficiency of KG incompleteness.
The explicit tree reasoning structure makes the results strongly interpretable.
Furthermore, we develop a new dataset called WikiPeopleQA to foster research on $n$-ary KGQA.
We then conduct comprehensive experiments on WikiPeopleQA dataset to show that \modelname~has the desired ability to perform effective reasoning on $n$-ary fact questions.
Besides, on two binary KGQA datasets, \modelname~also indicates a strong ability to infer answers compared with state-of-the-art methods accurately.

Our study fundamentally contributes to bridging the gap between binary KGQA and $n$-ary KGQA.
The proposed \modelname~can serve as a preliminary foundation for the $n$-ary KGQA. 
To summarize, our contributions are:

\begin{figure*}[t]
\centering
\includegraphics [width=0.99\textwidth]
{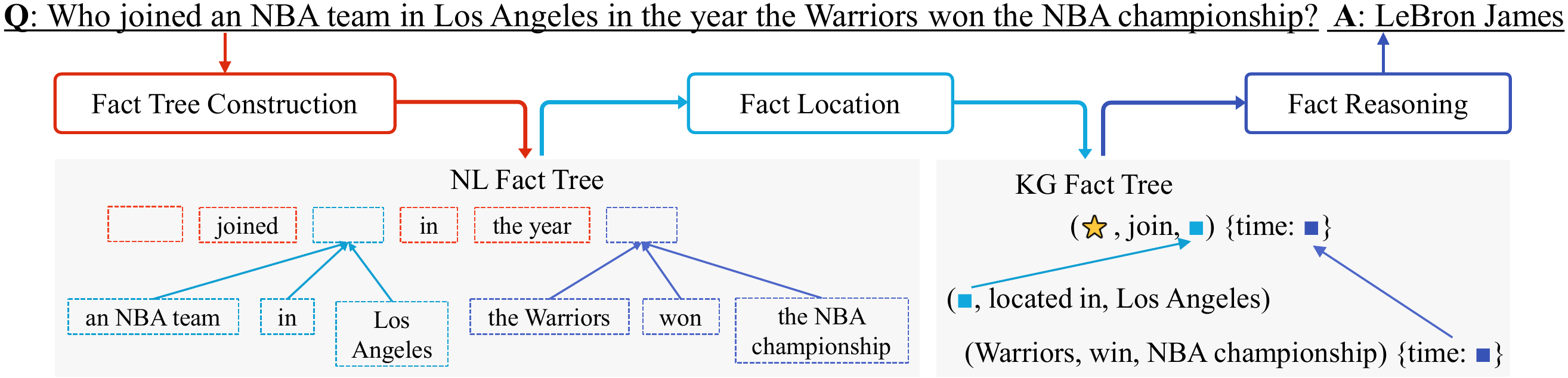}
\caption{
    Overview of \modelname. 
    It takes the question~(\textbf{Q}) as an input, passes through a three-stage pipeline processing: 1) fact tree construction~(Sec.~\ref{construction}), 2) fact location~(Sec.~\ref{distillation}) and 3) fact reasoning~(Sec.~\ref{reasoning}), and finally gets the answer entity~(\textbf{A}).
    Placeholders $\Box$ in the NL and KG fact trees indicate the entities to be inferred.
}
\label{fig:model}      
\end{figure*}

\begin{itemize}[leftmargin=*]
    \setlength{\itemsep}{-3pt}
    \setlength{\parsep}{-1pt}
\item[\tiny$\bullet$] 
We highlight a more challenging task: $n$-ary KGQA than a standard binary KGQA task setting. 
We further observe that the multi-hop reasoning framework popular in binary KGQA is no longer applicable to $n$-ary KGQA.

\item[\tiny$\bullet$] We propose a novel fact-tree reasoning framework, \modelname, which can serve as a preliminary foundation for $n$-ary KGQA study. And we develop a new dataset: WikiPeopleQA to foster research on $n$-ary KGQA. 

\item[\tiny$\bullet$] We conduct comprehensive experiments to show that our framework has the desired reasoning ability for both $n$-ary and binary KGQA tasks. 

\end{itemize}

\section{Related Work}
\label{relatedwork}

The previous series of KGQA models~\cite{liang2011learning,berant2013semantic,DBLP:conf/acl/YihHM14,lan-jiang-2020-query,sun2020sparqa,Tomer2020Break} synthesize a structured query graph from the question and then match the query with KG to get the answer.
This type of model has high interpretability but is challenged by the incomplete nature of KGs.
Then another series models compute the semantic similarity of the question and each candidate answer directly in the latent space~\cite{bordes2015large,dong2015question, Hamilton-2018-Embedding,zhang2018variational}.
This type of model overcomes the limitation of incomplete KG, but lacks sufficient interpretability.
\modelname~uses facts as the basic reasoning unit to alleviate the deficiency of KG incompleteness and the explicit tree reasoning structure to realize strongly interpretable.

Multi-hop reasoning framework has attracted widespread attention due to its high flexibility and high interpretability in recent years~\cite{fu2020survey}.
Current efforts build an explicit reasoning chain through training a reinforcement learning agent to walk on the KG~\cite{das2017go,Qiu2020Stepwise,kaiser2021reinforcement}, or construct implicit reasoning chains through memory network~\cite{NIPS2015_e2enn,miller2016key,chen2019bidirectional} or in the latent space~\cite{bordes-etal-2014-question,saxena2020improving,DBLP:conf/nips/RenL20,he2021improving,pmlrv139ren21a}.
This kind of method performs well on binary fact questions but 
has difficulties in dealing with $n$-ary fact questions.
Of course, one could construct and synthesize multiple reasoning chains to tackle $n$-ary KGQA.
But this would inevitably lead to an exponential increase in the reasoning difficulty and computational complexity, which in turn affects the reasoning performance.

Our work first highlights the $n$-ary KGQA task.
The research of $n$-ary KG provides a feasible research foundation for $n$-ary KGQA.
KG embedding learning on $n$-ary facts~\cite{wen2016representation,zhang2018scalable,fatemi2019knowledge,NaLP,guan2020neuinfer,Abboud2020BoxE} has grown considerably in recent years.
The $n$-ary KG embedding model~\cite{guan2020neuinfer} is plugged in \modelname~to alleviate the reasoning difficulties caused by the KG incompleteness.

\section{Fact-tree Reasoning}
\label{method}

\begin{figure*}[t]
    \centering
    \includegraphics [width=0.99\textwidth]
    {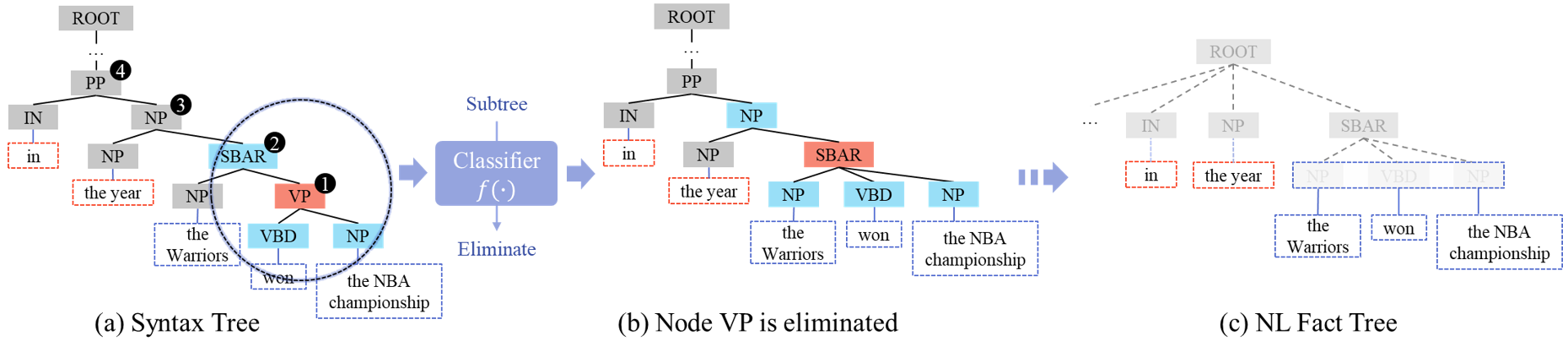}
    \caption{An example of NL fact tree construction~(partial steps). The eliminations of nodes in the order of \ding{182}$\rightarrow$\ding{183}$\rightarrow$\ding{184}$\rightarrow$\ding{185}$\rightarrow \cdots$.
    Best viewed in color.} 
    \label{fig:stage1}      
\end{figure*}

We illustrate the fact-tree reasoning framework for $n$-ary KGQA in Figure~\ref{fig:model}.
It takes the question~(\textbf{Q}) as an input, passes through a three-stage pipeline processing, and finally gets the answer entity~(\textbf{A}).
In the first fact tree construction stage, we construct the NL fact tree from the NL question~(Sec.~\ref{construction}).
In the second fact location stage, we locate the NL fact onto the KG to obtain the KG fact tree~(Sec.~\ref{distillation}). 
This helps to bridge the semantic and structure gap between the unstructured NL and the structured KG.
In the last fact reasoning stage, we perform intra-fact and inter-fact reasoning iteratively on the KG fact tree to infer the answer entity~(Sec.~\ref{reasoning}).

We use fact as the basic reasoning unit in the fact-tree reasoning framework.
In the NL fact tree, the fact is represented as a sequence of words and placeholders, where the words are taken from the NL question, and the placeholders refer to the missing entities to be inferred, e.g., the fact (\textit{$\Box$ joined $\Box$ in the year $\Box$}).
In the KG fact tree, following~\cite{guan2020neuinfer}, we represent the $n$-ary fact as 
\begin{equation}
fact\!=\!(s, p, o),\left \{a_{1}\!:\!v_{1},a_{2}\!:\!v_{2},...,a_{m}\!:\!v_{m}\right\}\!,
    \label{eq:fact}
\end{equation}
where $(s, p, o)$ denotes the subject-predicate-object information in the fact, named primary triple; each $\!a_{i}\!:\!v_{i}~(i\!\in\!\{1, 2, ... ,m\})$ is an attribute value pair, a.k.a., the auxiliary description to the primary triple.
For example, the ternary fact \textit{LeBron James joined Los Angeles Lakers in 2018} is formalized as \texttt{(LeBron~James, join,~Los~~Angeles~~Lakers)}, \texttt{\{time:} \texttt{2018\}}.
Note that the binary fact only contains the primary triple.

\subsection{Fact tree construction}
\label{construction}

In \modelname, we use fact as the basic reasoning unit and use the tree structure to represent the associations among facts.
Here, we design an automatic NL fact tree construction algorithm to transfer the NL question to the NL fact tree.
Since the syntax tree naturally expresses the hierarchical relation of the elements of the sentence and in order to facilitate the subsequent locating of the NL facts into KG, we use the syntax tree\footnote{We use the \href{https://nlp.stanford.edu/software/lex-parser.html}{Stanford Parser} to generate the syntax tree. The leaf nodes are words or phrases of {Q} and the branch nodes are syntax labels, e.g., NP~(Noun Phrase) and VP~(Verb Phrase).} of {Q} as the initial structure.
We expect the constructed NL fact tree to satisfy the following characteristics: 1) the leaf nodes are words or phrases of {Q}; and 2) if the leaf nodes share the same parent node, they belong to the same fact. 
Therefore, the NL fact tree construction algorithm can be viewed as an iterative eliminating of nodes in the syntax tree to achieve clustering of nodes within facts and differentiation between facts.

The node elimination process starts from the antepenult level of the syntax tree and proceeds from bottom to top.
We observe that two semantically different questions may be parsed into the same syntax structure except for the leaf nodes.
Therefore, disregarding leaf nodes makes our algorithm more adaptable.
Also, the parent node of a leaf node needs to be reserved for identifying the leaf node.
Figure~\ref{fig:stage1} shows a specific example
\footnote{Due to the space limitation, only part of the syntax tree and the corresponding elimination steps are shown here.}. As shown in (a), the pruning starts from the node VP~(\ding{182}, colored in red). To decide whether to eliminate this node or not, we extract a subtree that contains the node and its neighbor nodes~(colored in blue). This subtree is fed into a classifier $f(\cdot)$, which is composed of a Graph Convolutional Network~(GCN) as embedding layer and a fully-connected layer.
If $f(\cdot)$ outputs ``eliminate'', the node will be eliminated and its children will be directly connected to its parent, as shown in (b). Otherwise, this node will be retained. This process continues until the iteration meets the root node.
Finally, we remove non-leaf nodes and keep the hierarchical structure of leaf nodes. The nodes in the upper-layer facts that are connected to the lower-layer facts are replaced with placeholders, as shown in (c).
We summarize the construction of NL fact tree in Algorithm~\ref{algorithm:stage1}.



          \begin{algorithm}[t]
            \small
                \caption{\footnotesize{NL Fact Tree Construction}}
                \label{algorithm:stage1}
                \KwIn{The question \textbf{Q}, empty node stacks $\mathbb{V},\mathbb{V}'$;
                 }
              \KwOut{The NL fact tree FT;}
    
    
              \textbf{Initialization}\\ 
              \qquad$\rm FT = Parse(\text{\bf Q})$,\\ 
              \qquad $\mathbb{V}=$BFS(FT);\\
             \While  {$\mathbb{V}$}
              {
                $v=\mathbb{V}.pop()$;\\
                 \eIf{$v.isLeaf()$ \textbf{or} $v.children.isLeaf()$}{
                    \textbf{continue};
                } {$\mathbb{V}'.push(v)$;}
              }

    
    
              \While  {$v \neq \rm{FT}.root$}
              {
                $v=\mathbb{V}'.pop()$;
    
                $T_v=
                \left \{ v\right \}\cup \left \{v.parent\right \}\cup \left \{v.children \right \}$;
    
    
                \If {$f(T_v)==eliminate$}
                {
                    \textbf{update}\\
                    \For {each $child$ in $v.children$ }    {$child.set\_parent(v.parent, \text{FT})$}
                    FT$.delete(v)$
                }
    
    
           
              }
            \end{algorithm}

Specifically, for the GCN, we use the propagation rule for calculating the node embedding update for each layer as follows:
\begin{equation}
    \bm{h}_{v}^{(i+1)}=\sigma_{i}(\bm{h}_{v}^{(i)}\bm{W}_{0}^{(i)}+\sum_{u\in T_v\setminus v}\bm{h}_{u}^{(i)}\bm{W}_{1}^{(i)}),
\end{equation}
where $v$ and $T_v$ follow the definitions in Algorithm~\ref{algorithm:stage1}, $u$ is one of the neighbors of $v$, $\bm{h}_{*}^{(i)}$ represents the hidden layer activations of nodes in the $i^{th}$ layer, 
$\sigma_{i}(\cdot)$ is the activation function, and $\bm{W}_{0,1}^{(i)}$ are the $i$-layer weight matrices.

\subsection{Fact Location}
\label{distillation}

This stage aims to transfer the NL fact tree to the KG fact tree, specifically, to locate each NL fact in the tree to a KG fact.
It is divided into three specific steps: 1) entity linking, 2) structure matching, and 3) relation extraction.

During entity linking, following the standard setting in KGQA~\cite{saxena2020improving}, we assume that the entities of the question are given and linked to nodes on the KG.
Note that the placeholders are directly reserved and they indicate the entities that need to be inferred.

\begin{figure}
    \centering
    \includegraphics[width=0.43\textwidth]{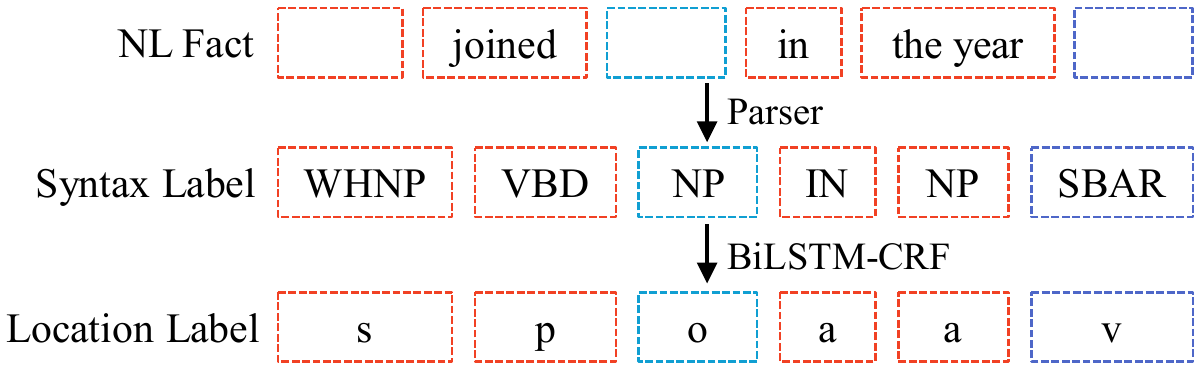}
    \caption{Example of a structure matching process.}
    \label{fig:tagging}
\end{figure}

The key of structure matching is to locate the subject $s$, predicate $p$, object $o$, attribute $a$ and value $v$ in the NL fact.
We view this process as a sequence labelling task, as shown in Figure~\ref{fig:tagging}.
We believe that location labels are more strongly associated with syntax labels, compared to word sequences.
Therefore, the input is the sequence formed by the syntax labels of each node in the NL fact.
The output is the sequence of location labels, and the label set is $\left \{ s,p,o,a,v \right \}$.
Here,we adopt the BiLSTM-CRF model~\cite{huang2015bidirectional} to perform sequence labelling.

After entity matching, we conduct relation extraction, i.e., transferring the word sequences labelled $p$ or $a$ to the corresponding relations (predicates or attributes) in KG.
Specifically, we adopt pre-trained SBERT model~\cite{reimers2019sentence} to get the embeddings of the relation in KG~(i.e., $\bm{r}$) and the word sequences~(i.e., $\bm{w}$). Then we use cosine similarity as a scoring function~$s(\cdot)$ to assign scores to the two embeddings and select the relation with the highest score:
\begin{equation}
    \begin{split}
        s(r,w)\!=\!\frac{r\cdot w}{\left \| r \right \| \!\left \| w \right \| } \!=\!\frac{ {\textstyle \sum_{i=1}^{n}\!r_{i}w_{i} } }{\sqrt{\sum_{i=1}^{n}\!r_{i}^{2} } \!\sqrt{\sum_{i=1}^{n}\!w_{i}^{2} }}.
    \end{split}
    \label{eq:relation}
\end{equation}

Finally, combining the predicted location labels, and linked entities and extracted relations, NL facts can be transformed into KG facts~(cf. Figure~\ref{fig:model}).
Placeholders are the bridge between upper-layer and lower-layer facts.
Interestingly, due to the variability of NL organization, there may be no placeholder in the lower-layer fact.
It is because when constructing the NL fact tree, the placeholder (usually value) is assigned to upper-layer fact according to the syntax structure. 
Therefore, we directly copy the upper-layer placeholder directly to the lower-layer fact.
For example, in Figure~\ref{fig:model}, the fact \textit{the Warriors won the NBA championship} will be transformed to \texttt{(Golden~State~Warriors, win, NBA} \texttt{championship)}, \texttt{\{time:2018\}}, where the attribute \texttt{time} is copied from the upper-layer fact.

\subsection{Fact Reasoning}
\label{reasoning}

\begin{figure}
    \centering
    \includegraphics[width=0.42\textwidth]{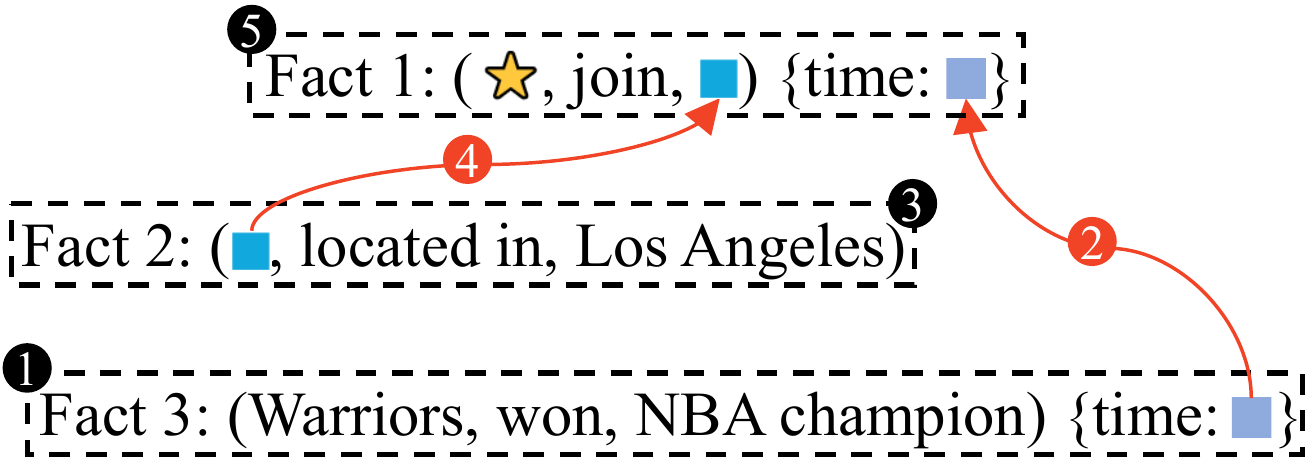}
    \caption{Example of an iterative reasoning process.
    The order of reasoning is \ding{182}$\rightarrow${\color{red0}\ding{183}}$\rightarrow$\ding{184}$\rightarrow${\color{red0}\ding{185}}$\rightarrow$\ding{186}.
    }
    \label{fig:resaoning}
\end{figure}

In this stage, we perform the inter-fact and intra-fact reasoning iteratively based on the KG fact tree to find the answer entity.
One example of iterative reasoning process is shown in Figure~\ref{fig:resaoning}.

\noindent
\textbf{Inter-fact Reasoning}
\label{interfact}
The whole process of inter-fact reasoning is carried out in a bottom-up manner.
Specifically, the entity inferred from the lower-layer fact will be transferred to the upper-layer fact.
For example, the entity \texttt{2018} inferred from fact 3 will be transferred to fact 1, i.e., the second step in Figure~\ref{fig:resaoning}.

    \begin{table*}[]
    
        \centering
        \caption{Statistics of WikiPeopleQA, WC2014 and PathQuestion, as well as their subsets. Note that the $n$-ary here means that $n$ is greater than $2$. }
        \label{tab:dataset}
        \resizebox{0.99\textwidth}{!}{
        \begin{tabular}{l|rrrrrr}\toprule[0.9pt]
        \multirow{3}{*}{Dataset } & \multicolumn{2}{c}{\# Question-Answer Pair} & \multicolumn{4}{c}{Background KG}\\  
        \cmidrule(lr){2-3}\cmidrule(lr){4-7}

        & \multicolumn{1}{c}{{Total}} & \multicolumn{1}{c}{{ Subset}}   &\# Fact   & \# Entity    & \# Relation(binary) & \# Relation($n$-ary)\\ \midrule[0.4pt] 
        WikiPeopleQA (1F/2F/3F)             & 4,491                 & 2,365    / 1,497   / 629~~     &  56,426            & 28,043   &  150~~~~~~~~~ & 557~~~~~~~~~           \\ 
        WC2014 (1H/2H/C)     & 10,162   & 6,482 / 1,472 / 2,208      & 6,482         & 1,127             &  6~~~~~~~~~ & 0~~~~~~~~~      \\ 
        PathQuestion (2H/3H) & 7,106 &1,908 / 5,198 & 4,049 & 2,215 & 14~~~~~~~~~ & 0~~~~~~~~~ \\ 
        \midrule[0.9pt]  
        \end{tabular}}
        \end{table*}

\noindent
\textbf{Intra-fact reasoning}
\label{intrafact}
This module aims to infer the missing entity of each incomplete fact.
We formulate this process as the KG completion task.
KG embedding models~\cite{bordes2013translating,Dettmers2018Convolutional,NaLP,guan2020neuinfer} are studied to deal with this task, by learning entity and relation embeddings and designing a scoring function to infer the missing entity.
In this work, we use NeuInfer~\cite{guan2020neuinfer}, a KG embedding model that can perform on binary and $n$-ary facts, to implement intra-fact reasoning.

However, comparing with the traditional KG completion task, the missing entity needs not only to complete the current fact, but also to satisfy the upper-layer fact. 
For example, in fact 2, \texttt{Sunset~Boulevard} and \texttt{Los Angeles Lakers} are all located in Los~Angeles. While, considering the upper-layer fact 1, the missing entity needs to satisfy the fact that the predicate is \texttt{join}.
Therefore, we introduce a score amplification mechanism:
if an alternative entity can satisfy the upper-layer fact, its corresponding score will be magnified $\lambda$ times.

\subsection{Training}

The classifier~$f(\cdot)$ and BiLSTM-CRF model are trained in a supervised manner,
where the training signals are obtained from manually labeled (syntax tree, NL fact tree) and (syntax label sequence, location label sequence)  pairs, respectively.
We observe the syntax structure of different questions may be similar or even consistent.
Therefore, we reduce the the input space by using syntax-related information rather than NL, to relieve the manual annotation pressure and also the learning difficult.

\section{Experiments}
\label{experiments}

\subsection{Dataset}
\label{dataset}

In this work, we target at studying the $n$-ary KGQA task.
Considering the popular KGQA datasets involve almost exclusively binary facts, we develop an $n$-ary KGQA dataset: WikiPeopleQA~(abbr. WP), in which questions involve multiple $n$-ary facts and the background KG is also composed of $n$-ary facts.
We also conduct evaluation on two binary KGQA benchmarks: WC2014~(abbr. WC)~\cite{zhang2016gaussian} and PathQuestion~(abbr. PQ)~\cite{zhou2018interpretable}.
Depending on the number of \textbf{F}acts or \textbf{H}ops involved in the question\footnote{Note that, for a binary fact question, the hop number is generally equal to the fact number.}, WikiPeopleQA is divided into WP-1F, WP-2F and WP-3F.
WC2014 is divided into WC-1H and WC-2H, as well as a conjunctive question set WC-C.
PathQuestion is divided into PQ-2H and PQ-3H.
We partition the three datasets into train/valid/test subsets with a proportion of 8 : 1 : 1.
The detailed statistics are shown in Table~\ref{tab:dataset}.

\subsection{Experimental Setup}
\label{setup}

\subsubsection{Training Details}
\label{training1}
During training the classifier, the embedding size of node is 50, the learning rate is $1e{-5}$, and the number of GCN layers is 3. 
During training the BiLSTM-CRF model, the embedding size of node is 100, the learning rate is $2e{-5}$. 
We use Adam optimizer for optimization for above training process.
The hyper-parameter $\lambda$ is set to 1.5.
The training process of the KG embedding model used in the fact reasoning stage is following~\cite{guan2020neuinfer}.
The experimental results are all averaged across three training repetitions.

\begin{table*}[t]
    
    \centering
    \Huge
    \renewcommand{\arraystretch}{1.23}
    \resizebox{1\textwidth}{!}{
    \begin{tabular}{l|cccc|cccc|ccc}
        \toprule[2.5pt]

        Model & WP & WP-1F & WP-2F & WP-3F & WC & WC-1H & WC-2H & WC-C  & PQ & PQ-2H & PQ-3H \\\midrule[1.5pt]
    
    MemNN~\cite{NIPS2015_e2enn}     & \underline{32.9}   & 34.2  & 39.6 & 12.5   & 52.4            & 71.6           & 55.5   &  73.3     & 86.8                & 89.5         & 79.2          \\
    KV-MemNN~\cite{miller2016key}    &   24.5      &  15.0      &    \underline{40.0}     & \underline{16.0}  & 76.7             & 87.0        &   87.0   &    78.8  &   85.2    & 91.5      & 79.4                      \\
    EmbedKGQA~\cite{saxena2020improving} & 26.4 & 35.4 & 22.3 & 3.5  &  52.5       &  59.6     & 79.0   &52.0  &  36.7           & 51.0         &  30.6                 \\
    IRN-weak~\cite{zhou2018interpretable}     &     -         &          -   &   -        & -  & 78.6         & 83.4       &  92.1   & 83.7 & 85.8     & 91.9     & 83.3                \\
    MINERVA~\cite{das2017go}     &     10.9         &     20.5        &      0.3     & 0.2& 89.6     &  87.2      & 93.1      &  82.4    & 73.1           &   75.9   & 71.2       \\
    SRN~\cite{Qiu2020Stepwise}      & 13.3     &  24.9       &  0.3     & 0.8  & \underline{96.5}             & \underline{98.9}         & \textbf{97.8}     & \underline{87.3}   &\underline{89.3}              &\underline{96.3}            &\underline{89.2}              \\
    QGG~\cite{lan-jiang-2020-query}     & 24.9    & \underline{41.8}  & 1.1 & 0.0   & 94.0            & 94.9           & 92.9   &  \textbf{99.9}     & 40.4                & 67.8         & 28.9          \\
    \midrule[1.5pt]
    \modelname    & \textbf{54.4} & \textbf{63.1} & \textbf{47.0} & \textbf{40.1}  & \textbf{99.5}       &  \textbf{99.9}      & \underline{96.3} & \textbf{99.9}  &  \textbf{92.8}      &  \textbf{98.4}      & \textbf{90.8}      \\ 
    \midrule[2.5pt] 
    \end{tabular}}
    \caption{Model Performance on $n$-ary KGQA task~(WikiPeopleQA dataset) and binary KGQA task~(WC2014 and PathQuestion datasets) under the accuracy(\%) metric~(pairwise t-test at 5\% significance level).
    The best performance results are shown in \textbf{bold}, and the second best results are shown in \underline{underlined}.
    }
    \label{tab:main}

    \end{table*}

\subsubsection{Baselines}
\label{baseline}

We compare our framework with a series of popular multi-hop reasoning baseline approaches.
According to the form of the reasoning chain, these baselines can be divided into two categories.
The first category baseline builds an explicit reasoning chain through training an agent to walk on the KG: IRN-weak~\cite{zhou2018interpretable}, MINERVA~\cite{das2017go} and SRN~\cite{Qiu2020Stepwise};
the second category baseline builds an implicit reasoning chain through memory network or in the latent space: MemNN~\cite{NIPS2015_e2enn}, KV-MemNN~\cite{miller2016key} and EmbedKGQA~\cite{saxena2020improving}.
We also compare with QGG~\cite{lan-jiang-2020-query}, which synthesizes a query graph from the question and then match the query with KG to get the answer.
When evaluating explicit multi-hop reasoning methods and QGG, we use the dummy entity to divide $n$-ary facts in the KG of WikiPeopleQA into binary facts.

\subsection{Main Results}
\label{exp1}

\noindent
\textbf{Performance of \modelname}
Table~\ref{tab:main} presents the statistics of model’s performances both on the $n$-ary and binary KGQA tasks.
We can see that \modelname~achieves significantly higher accuracy than state-of-the-art baselines on the $n$-ary KGQA task.
Specifically, compared with the best performing multi-hop reasoning baseline MemNN, \modelname~improves accuracy by 21.5\%~(w.r.t., WP).
We have following discoveries:

\begin{itemize}[leftmargin=*]
    \setlength{\itemsep}{0pt}
    \setlength{\parsep}{-1pt}
    \item[\tiny$\bullet$] \modelname~shows large advantages on coping with complex questions with multiple facts. On the WP-3F sub-dataset, comparing with baselines, \modelname~has made a qualitative leap on accuracy~(16.0$\rightarrow$40.1).
    This confirms that multi-hop reasoning methods are inapplicable to more complex reasoning scenarios.

\item[\tiny$\bullet$] Interestingly, the explicit multi-hop reasoning methods~(e.g., SRN) are obviously weaker than implicit methods~(e.g., EmbedKGQA). This is because the $n$-ary facts are split by dummy entities, adding difficulty to build explicit reasoning chains.
Implicit methods weaken the distinction between binary and $n$-ary facts, which makes it more flexible in dealing with $n$-ary facts.

\item[\tiny$\bullet$] The QGG method directly matches the generated query graph to the KG, resulting in performing well on questions with less facts, but clearly lacking flexibility for question with more facts.

\end{itemize}

Besides, fact-tree reasoning also achieves a good performance on binary KGQA.
We observe a large performance gap between multi-hop reasoning methods on binary and $n$-ary KGQA tasks.
Therefore, it would be valuable to pay more attention to the study of $n$-ary KGQA.

\begin{figure}
    \centering
    \includegraphics[width=0.4\textwidth]{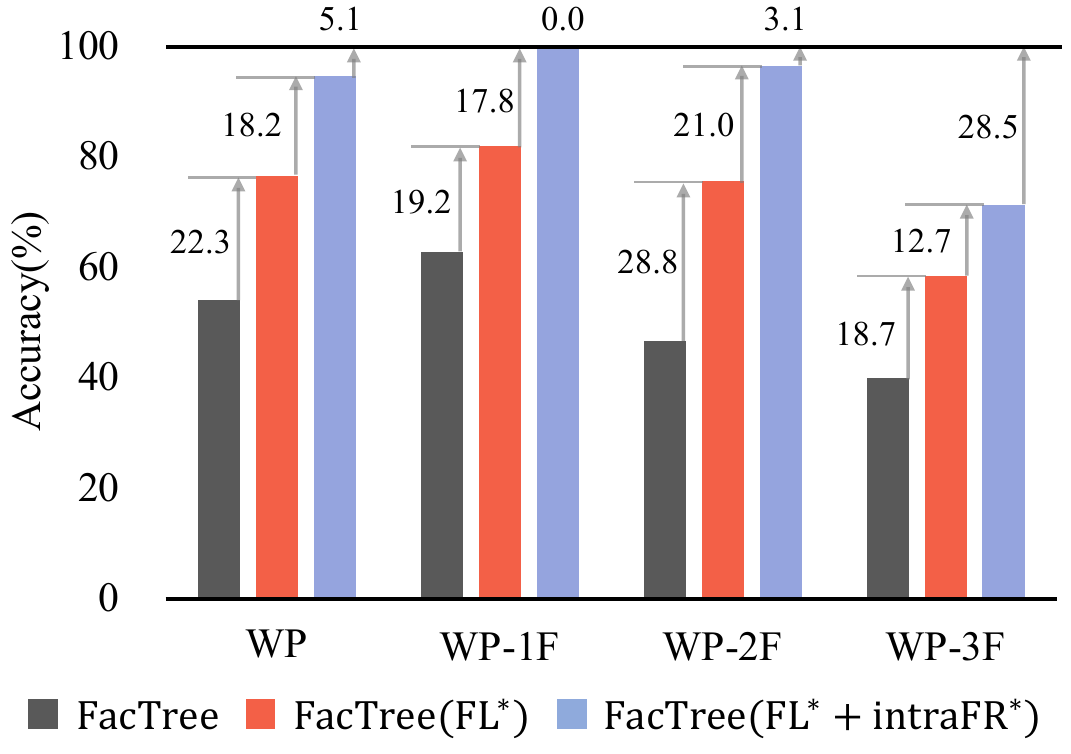}
    \caption{Staged evaluation of \modelname{}. $\rm{FL}^*$ and $\rm{intraFR}^*$ denote turning off the FL and intraFR components respectively. Note that the accuracy of \modelname{}($\rm{FL}^*$+$\rm{intraFR}^*$+$\rm{interFR}^*$) is 100\%.}
    \label{fig:ablation}
\end{figure}

\begin{figure*}[t]
\centering
\includegraphics [width=1\textwidth]
{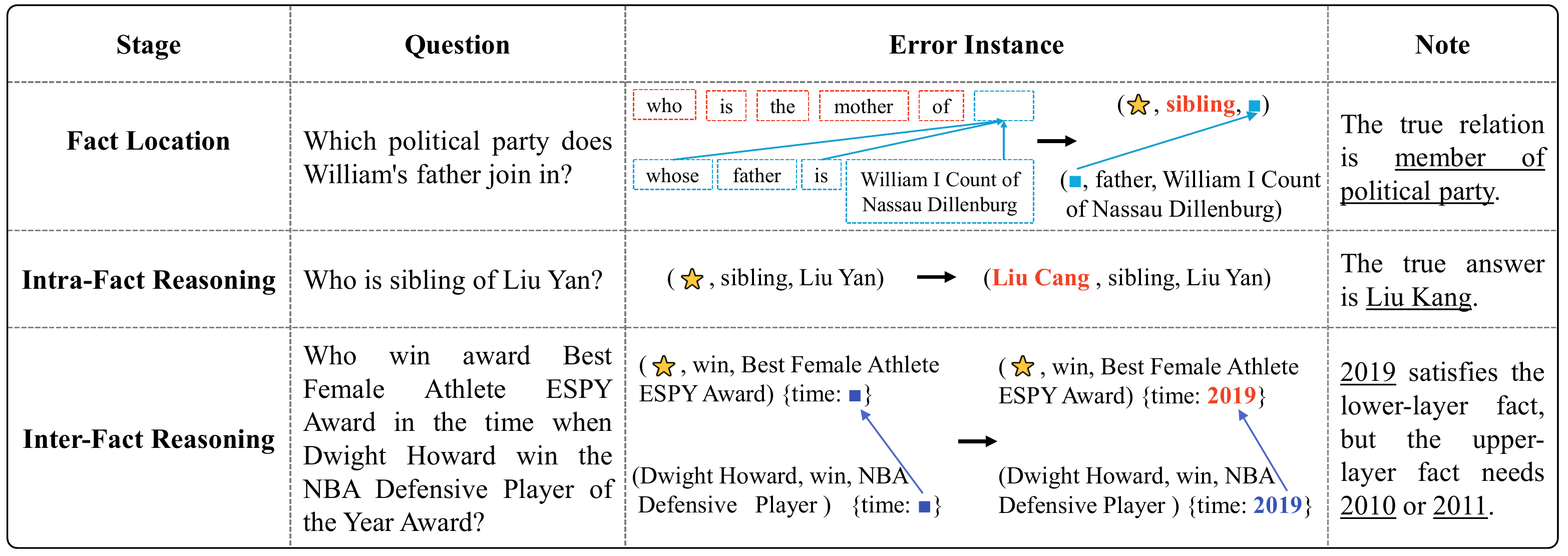}
\caption{
   Error instances in the fact location, intra-fact and inter-fact reasoning stages, respectively.
}
\label{fig:case}      
\end{figure*}

\noindent
\textbf{Staged Evaluation}
We test the capability of each stage of \modelname.
Because the fact-tree reasoning framework is a pipeline structure, the reasoning error always occurs in cascade.
We turn off the components on the pipeline from the beginning to see the impact on the overall effect.
``Turn off a component'' means to replace the real output with the ground truth, that is, the component is perfect by default.
We test three components of \modelname: fact location~(FL), intra-fact reasoning~(intraFR) and inter-fact reasoning~(interFR).
The error caused by the fact tree construction stage is negligible here because of the relatively small number of NL fact tree types in the datasets.
Figure~\ref{fig:ablation} shows turning off the components in turn will lead to an accuracy increase of 22.3\% (FL), 18.2\% (intraFR) and 5.1\% (interFR) respectively~(w.r.t., WP).
Impressively, transferring the NL fact tree to KG fact tree and inferring missing entities of incomplete facts are the two keys that affect the reasoning accuracy.
Moreover, the influence of the inter-fact reasoning module becomes apparent as the number of facts increases.
This is because it may happen that an entity satisfying the lower-layer fact may not be able to satisfy the upper-layer fact.
For instances of errors in each module, please see Section~\ref{exp2}.

\subsection{Analysis of \modelname}
\label{exp2}

\noindent
\textbf{Performance w.r.t. Incomplete KGs}
As the capacity of the KG continues to expand, current KGs are typically incomplete with many facts missing.
Incomplete KGs put forward higher requirements on the capabilities of KGQA models.
Therefore, we conduct an experiment on \modelname~and two popular baselines to test their reasoning capability for incomplete KGs.
As shown in Table \ref{tab:gcn}, when the KG is reduced by half, the effect of our model decreases the least.
This is because we adopt KG embedding models to perform intra-fact reasoning in \modelname.
This design relaxes the requirements for KG completeness.
SRN requires the construction of query graph or explicit reasoning chain on KG, so it is more sensitive to the incompleteness of KG.
EmbedKGQA also uses KG embedding models.
The performance gap between it and our \modelname~corroborates the superiority of the fact-tree reasoning framework.

\noindent
\textbf{Zero-shot Learning w.r.t. Classifier~$f(\cdot)$}
We conduct a zero-shot learning experiment to test the capability of our proposed classifier~$f(\cdot)$ in the fact construction stage though a 5-fold cross validation.
We evaluate the accuracy of the constructed NL fact tree on the mixed dataset combined with three datasets~(WikiPeopleQA, WC2014 and PathQuestion).
The mixed dataset is divided into five parts according to the NL fact tree classes.
For each fold, the fact tree classes in the testing set do not appear in the training set.
Based on this setting, our classifier can reach 81.2\% accuracy with a standard deviation of 0.098.
This indicates our classifier has the scalable ability to construct unseen fact trees.

\begin{table}[t]
    \centering
        \resizebox{0.49\textwidth}{!}{
            \begin{tabular}{l|cc|cc}
                \toprule[1.2pt]
                Model &WP&WP-50\%&WC&WC-50\%\\\midrule[1.0pt] 
                    SRN& 13.3 & 0.1 {\small ($\downarrow$99\%)} & 96.5 & 0.0 {\small ($\downarrow$100\%)}\\
                EmbedKGQA&26.4 &6.5 {\small ($\downarrow$75\%)}&52.5 &11.0 {\small ($\downarrow$79\%)}\\
                \modelname &54.4 &17.8 {\small ($\downarrow$67\%)}&99.5 &37.2 {\small ($\downarrow$63\%)}\\
                \midrule[1.2pt] 
            \end{tabular}
        }
        \caption{Performance on incomplete KGs.
        $\downarrow$ indicates the decrease in accuracy when the KG is halved.
        }
        \label{tab:gcn}
    \end{table}

\noindent
\textbf{Effectiveness of GCN}
We also test the effect of the execution range of the GCN on the accuracy of the fact tree construction.
The range of GCN execution is a subtree of the syntax tree containing the central node to be eliminated and its neighbor nodes.
A total of five range types are tested depending on whether the father, siblings and child of the central node were included.
As shown in Table~\ref{tab:treesize}, the optimal subtree range includes the central node and its father and child nodes.
Interestingly, the addition of sibling nodes did not bring significant effect improvement.

\begin{table}[t]
    \centering
        \normalsize
        \resizebox{0.40\textwidth}{!}{
            \begin{tabular}{l|ccccc}
            \toprule[1.5pt]
               Size&O&O+F&O+C&O+F+C&O+F+C+S\\\midrule[1.0pt] 
              Acc. &52.9&53.9& 63.1 &91.4& 91.0\\
            \midrule[1.5pt] 
            \end{tabular}
        }
        \caption{Performance of different subtree range w.r.t. fact tree construction stage.
    ``F'', ``C'' and ``S'' denote the father, child and sibling nodes of the central node ``O'', respectively.
    }
        \label{tab:treesize}
    \end{table}

\noindent
\textbf{Error Analysis}
We conduct a qualitative study on error instances, as shown in Figure~\ref{fig:case}, and analysis the directions for future work.  
In the fact location stage, introducing more effective relation extraction techniques can contribute to reduce relation extraction error.
In the intra-fact reasoning stage, it is necessary to improve KG embedding model's capability.
There is a false negative error case.
For example, the fact \texttt{(Liu Cang, sibling, Liu Yan)} is true, but is not included by KG, resulting in the inferred answer being judged as wrong.
So, adopting broader KGs is suggested in future studies.
In the inter-fact reasoning stage, entities can be incompatible when transferred between facts.
Therefore, intra- and inter-factual reasoning needs to act more closely together to reduce the incompatibility.

\vspace{2pt}
\section{Conclusion}
\label{exp3}

\vspace{6pt}

This work highlights a more challenging task: $n$-ary KGQA, and it advocates that the multi-hop reasoning framework popular in binary KGQA is no longer applicable to $n$-ary KGQA.
A novel fact-tree reasoning framework \modelname~is proposed, which pipelines the $n$-ary KGQA into three steps: fact tree construction, fact location, and fact reasoning to infer the correct answer.
The quantitative and qualitative experimental results have demonstrated that \modelname~has superior reasoning ability on $n$-ary and binary fact questions.

\section{Acknowledgements} 
This work was supported in part by National Natural Science Foundation of China under Grant No.62006101, and in part by the China Postdoctoral Science Foundation under Grant No.2021TQ0222 and No.2021M700094.

\bibliography{anthology,custom,reference}
\bibliographystyle{acl_natbib}

\appendix
\appendix

\section{Methodology Details}

\subsection{NL Fact Tree Construction Algorithm}
\label{sec:algorithm}

The construction of NL fact tree is summarized in Algorithm~\ref{algorithm:stage1}. 
The input is the NL question and two empty node stacks. The output is the NL fact tree. We initialize the NL fact tree as a syntax tree, which is parsed from \textbf{Q}~(Line 2).
One of the empty node stack $\mathbb{V}$ stores the nodes of FT in the order of breadth first searching~(Line 3). The other stack $\mathbb{V}'$ reverse this order in Line 4-10, so that the pruning~(Line 12-22) is from the the bottom to the top as well as from the right to the left.
The entire elimination process does not involve the leaf nodes and their parents nodes~(Line 6-10).

\subsection{Score Amplification Mechanism}

We devise a simple method to evaluate whether an entity is able to satisfy the upper-layer fact.
For an entity $e$, the related predicate or attribute in the upper-layer fact is $p$ or $a$.
We retrieve in KG, whether there is an fact of entity $e$ associated with $p$ or $a$.
If there is, then we consider that entity $e$ is satisfying the upper-layer fact, and vice versa.

\section{Experimental Details}

\subsection{Dataset Construction}

In this work, we develop an $n$-ary KGQA dataset: WikiPeopleQA, in which questions involve multiple $n$-ary facts and the background KG is also composed of $n$-ary facts.
The specific construction process is as follows:

\begin{figure}[t]
    \centering
    \includegraphics[width=0.42\textwidth]{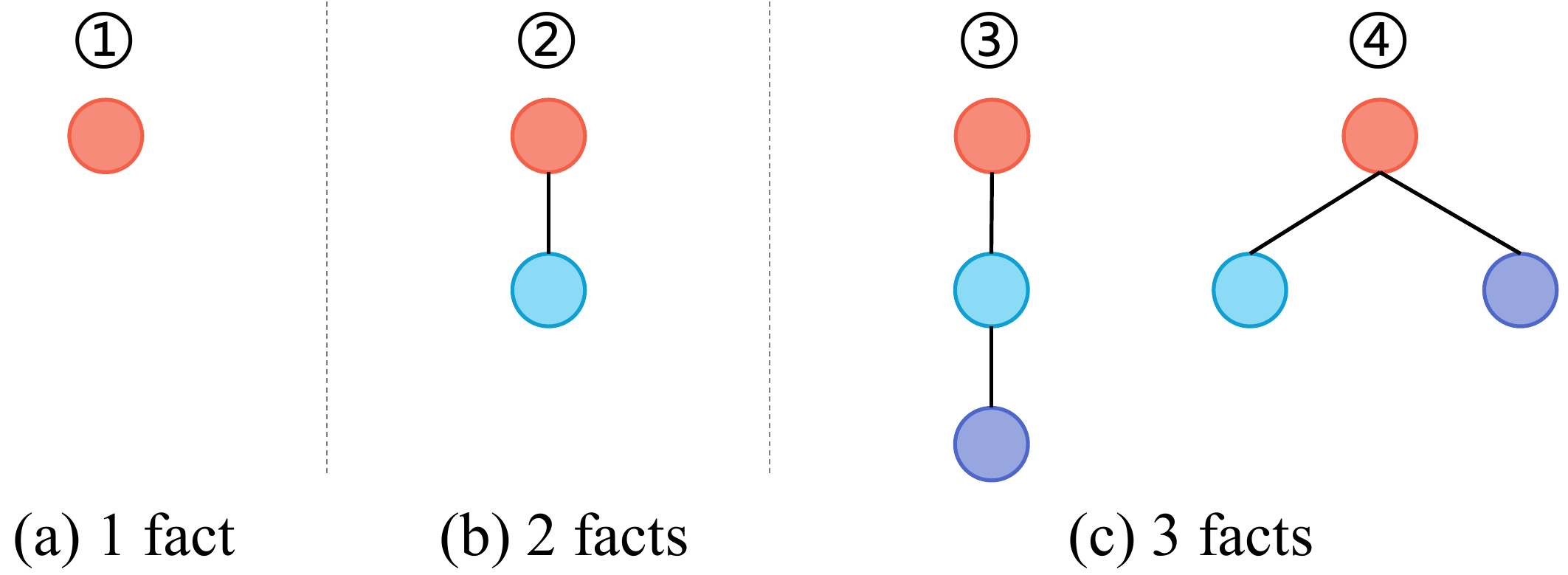}
    \caption{The fact combination modes.}
    \label{fig:factmode}
\end{figure}

\begin{itemize}

    \item[i)] We selected~WikiPeople~\cite{NaLP} as the background KG.
    This $n$-ary KG is constructed based on Wikidata\footnote{\url{https://www.wikidata.org}} and consists of character facts, e.g., \textit{Marie Curie received Nobel Prize in Chemistry in 1911}.

    \item[ii)] To build complex questions involving multiple facts, we set the maximum number of facts in a question to three, and set four fact combinations modes in advance, as shown in Figure~\ref{fig:factmode}.
    As the number of facts increases, the fact combination mode becomes more complicated.

    \item[iii)] Based on the fact combination modes, we sampled a large number of fact combinations from KG. We masked off the entities in the fact combinations, and extracted the frequently occurring fact combinations. We transformed the frequent fact combinations to question templates.
    Here we have constructed a total of 33 question templates, listed in Table~\ref{tab:questiontemplate}.

    \item[iv)] We populated the entities into the question templates according to KG. Inspired by the construction process of PathQuestion~\cite{zhou2018interpretable}, in order to enrich the problematic syntactic structure and surface wording, we replaced the phrases and words in the question with synonyms.

\end{itemize}

Due to the limitation of fact diversity in WikiPeople, we only considered three facts at most. 
In order to contribute to the progress of $n$-ary KGQA research, it is necessary to increase the richness of the facts to improve the question complexity. Therefore, developing more complex $n$-ary KGQA datasets and evaluating \modelname{} on more datasets are our future research directions.

\subsection{Baselines}

We compare our framework with a series of baselines. 
The following is a detail description of the baselines:

\begin{itemize}[leftmargin=*]
    \setlength{\itemsep}{0pt}
    \setlength{\parsep}{0pt}
    \item[\tiny$\bullet$] MemNN~\cite{NIPS2015_e2enn}: This model adopts an memory network to store all KG facts or related Wikipedia documents in the memory units. Three embedding matrices are employed to convert the memory information and questions into vectors for similarity calculation.
    \item[\tiny$\bullet$] KV-MemNN~\cite{miller2016key}: This model is based on MemNN. Instead of considering the whole KG facts like MemNN, it firstly stores facts in a key-value structured memory. The key-value structure is suitable for binary facts, but not for $n$-ary facts.
    \item[\tiny$\bullet$] EmbedKGQA~\cite{saxena2020improving}: This model follows the basic multi-hop reasoning framework and utilizes KG embedding methods to alleviate the negative impact of KG incompleteness.     
    \item[\tiny$\bullet$] IRN-weak~\cite{zhou2018interpretable}: This model considers the whole path from the topic entity to the answer entity. It focuses on finding a path to the answer, so IRN needs a pre-labelled path record during training process.
    \item[\tiny$\bullet$] MINERVA~\cite{das2017go}: This model uses reinforcement learning technique to perform multi-hop reasoning on KG. Taking the input natural language question, this model averages the word embeddings as the question embedding, and then walks on KG under the supervision of the question embedding, and finally arrives at the answer entity.
    \item[\tiny$\bullet$] SRN~\cite{Qiu2020Stepwise}: This model uses RL method to perform multi-hop reasoning on KG. It proposes a potential-based reward shaping strategy to alleviate the delayed and sparse reward problem caused by weak supervision.
    \item[\tiny$\bullet$] QGG~\cite{lan-jiang-2020-query}: This model generates a modified staged query graph to deal with complex questions with both multi-hop relations and constraints.
    
\end{itemize}

Here we explain the source of the results in Table~\ref{tab:main}.
On the WikiPeopleQA dataset, for each baseline, we run the source code of each baseline that is open source or reproduced by developers.

\begin{itemize}[leftmargin=*]
    \setlength{\itemsep}{0pt}
    \setlength{\parsep}{0pt}
    \item[\tiny$\bullet$] MemNN: \url{https://github.com/berlino/MemNN} (reproduced)
    \item[\tiny$\bullet$] KV-MemNN: \url{https://github.com/lc222/key-value-MemNN}~(reproduced)
    \item[\tiny$\bullet$] EmbedKGQA: \url{https://github.com/malllabiisc/EmbedKGQA}~(open source)
    \item[\tiny$\bullet$] MINERVA: \url{https://github.com/shehzaadzd/MINERVA}~(open source)
    \item[\tiny$\bullet$] SRN: \url{https://github.com/DanSeb1295/multi-relation-QA-over-KG}~(reproduced)
    \item[\tiny$\bullet$] QGG:~\url{https://github.com/lanyunshi/Multi-hopComplexKBQA}~(open source)
    
\end{itemize}

IRN-weak needs the pre-labelled path records for training, which is not applicable to the $n$-ary KGQA task.
So we do not evaluate IRN-weak.

For two binary KGQA datasets WC2014 and PathQuestion, the results of EmbedKGQA and QGG are obtained by our own tests.
Other baseline results all cited from~\cite{Qiu2020Stepwise}.

\subsection{Training Details}

Note that in the NL fact tree, there is overlap between facts. 
When we construct the (syntax tree, NL fact tree) training samples, the overlap will be maintained in the lower-layer fact if it belongs to the primary triple, and in the upper-layer fact if the overlap belongs to the auxiliary description.
For example, in Figure~\ref{fig:model}, 
the overlapping part of fact 1 and fact 2 ``an NBA team'' belongs to the primary triple, so it is maintained in fact 2.
Conversely, ``in the year'' is the auxiliary description and is maintained in fact 3.

\section{Example of NL Fact Tree Construction}
\label{sec:treeexample}

Here we display a visual example of the NL fact tree construction with the question \textit{Who joined an NBA team in Los Angeles in the year the Warriors won the NBA championship}.

\begin{figure*}
    \centering
    \includegraphics[width=0.89\textwidth]{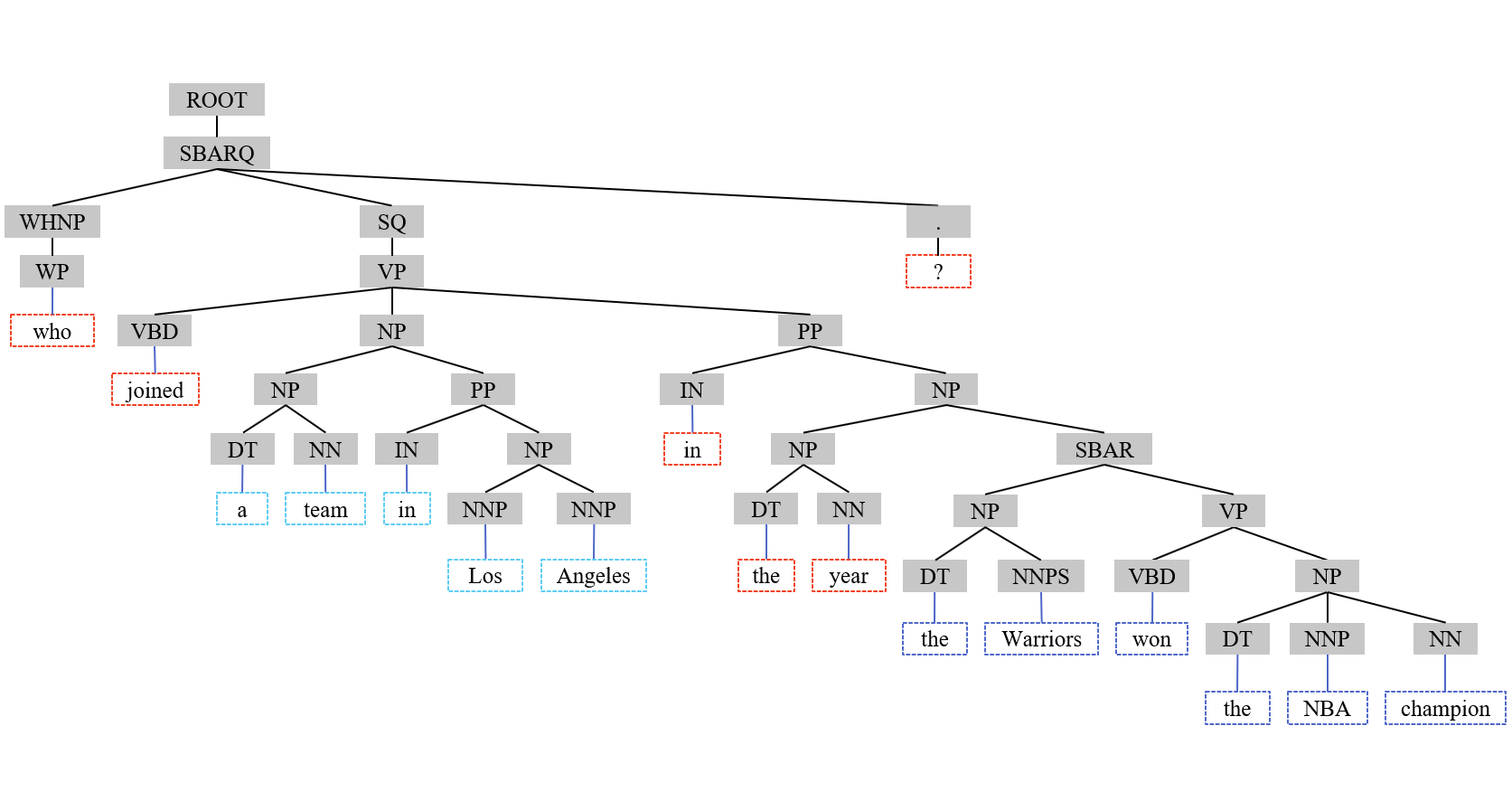}
    \caption{Syntax tree.}
    \label{fig:1}
    
\end{figure*}

\begin{figure*}
    \vspace{-3cm}
    \centering
    \includegraphics[width=0.89\textwidth]{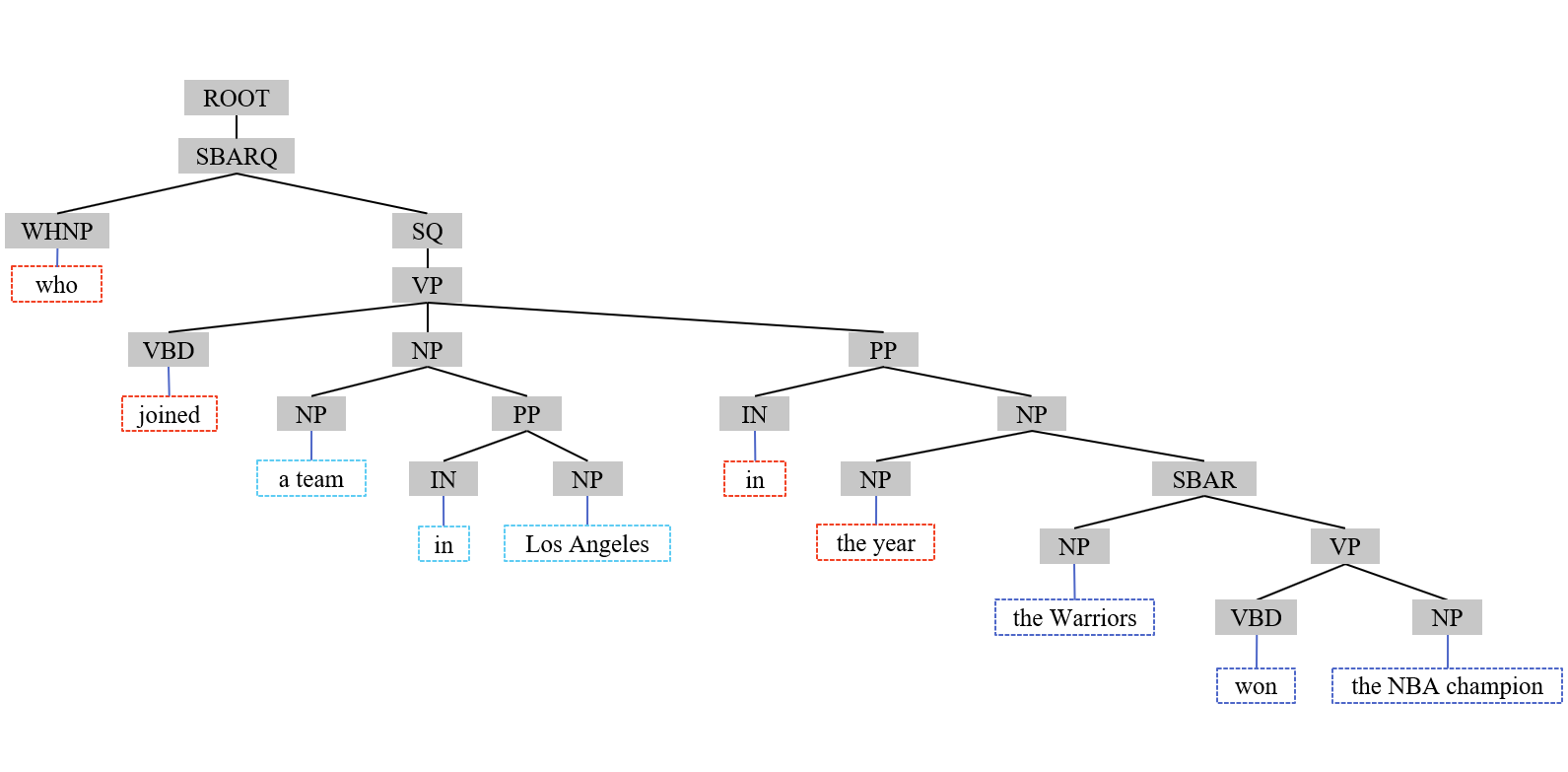}
    \caption{Syntax tree after preprocessing.}
    \label{fig:2}
    \vspace{-3cm}
\end{figure*}

Firstly, we use the Stanford Parser to generate the syntax tree~(cf. Figure~\ref{fig:1}).
Then we preprocess the syntax tree~(cf. Figure~\ref{fig:2}) to reduce the subsequent elimination operations according to the following rules:

\begin{itemize}[leftmargin=*]
    \setlength{\itemsep}{0pt}
    \setlength{\parsep}{0pt}
    \item[\tiny$\bullet$] Pruning the punctuation node and its parent node, e.g., node ``?'' and ``.''.

    \item[\tiny$\bullet$] If all the grandchildren of a NP node are leaf nodes (more than one), 
    we prune the parents, and combine the grandchildren to a unified leaf node, whose parent is changed to the NP node. 
    For example, \texttt{the}, \texttt{NBA} and \texttt{championship} in Figure~\ref{fig:1} are combined as \texttt{the NBA championship}, whose parent is ``NP''.

\item[\tiny$\bullet$] If a node has only one child and only one grandchild, which is a leaf node, we remove the child node and let the leaf node be the only child of this node. For example, \texttt{who} in Figure~\ref{fig:1} will be connected directly to its grandfather node \texttt{WHNP}.

\end{itemize}

\begin{figure*}[htbp]
    \centering
    
    \subfigure[VP is eliminated.]{
    \begin{minipage}[t]{0.49\linewidth}
    \centering
    \includegraphics[width=0.99\textwidth]{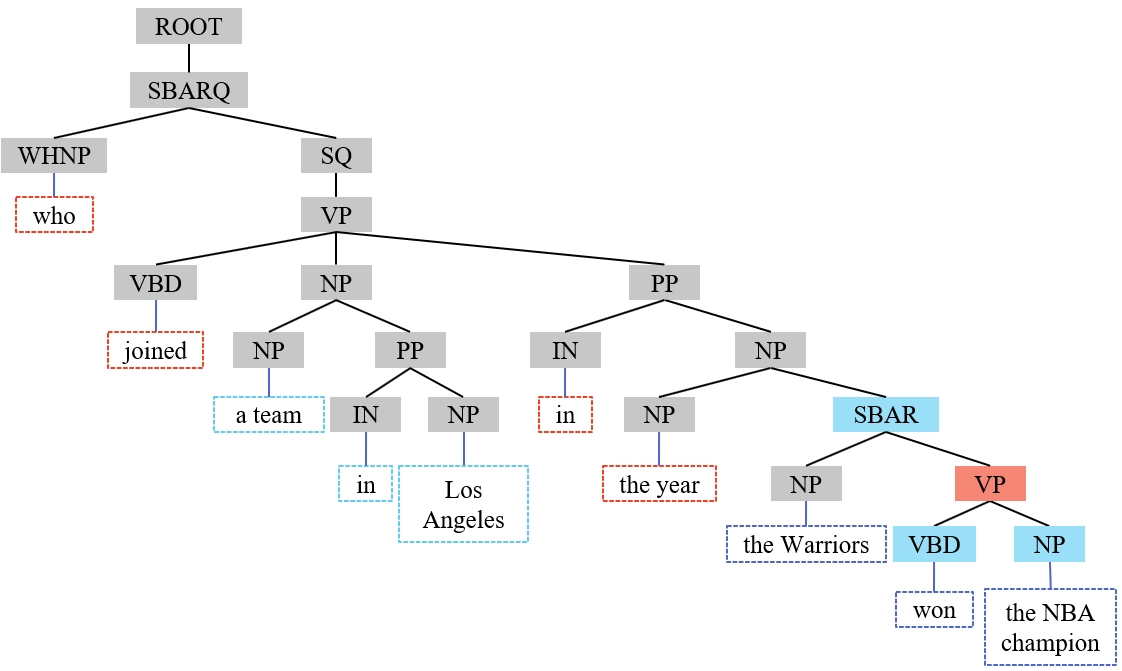}
    \end{minipage}%
    }%
    \subfigure[SBAR is retained.]{
    \begin{minipage}[t]{0.49\linewidth}
    \centering
    \includegraphics[width=0.99\textwidth]{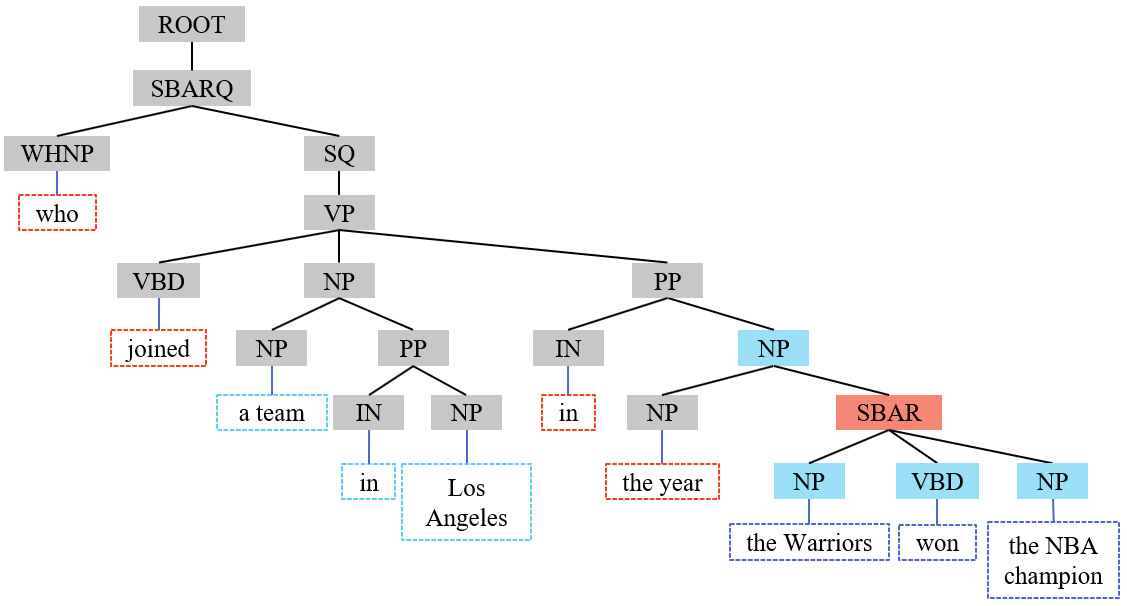}
    \end{minipage}%
    }%

    \subfigure[NP is eliminated.]{
        \begin{minipage}[t]{0.49\linewidth}
        \centering
        \includegraphics[width=0.99\textwidth]{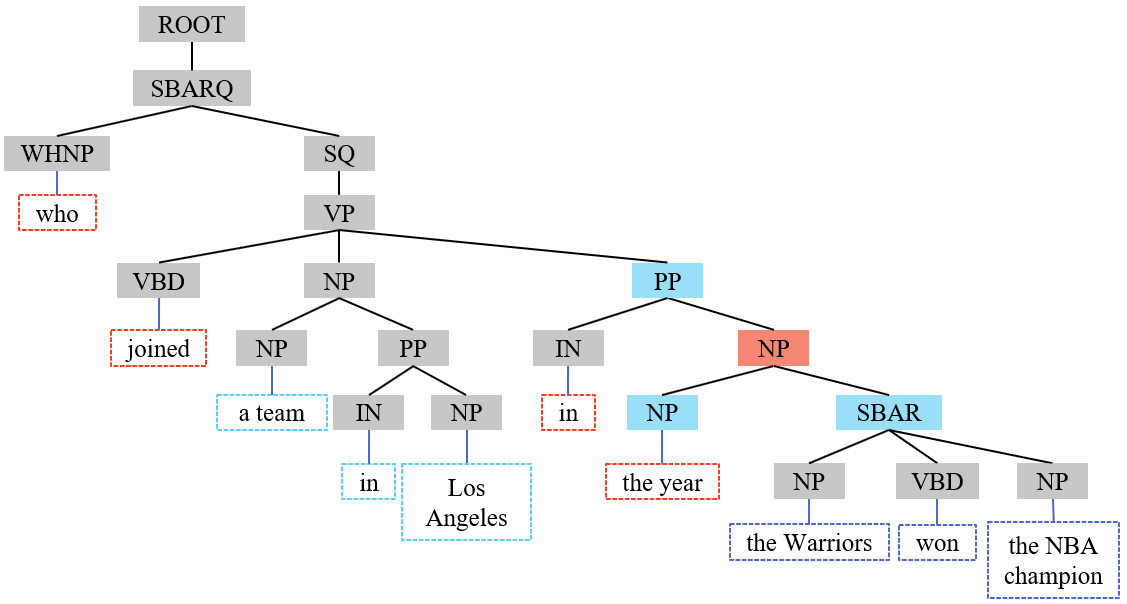}
        \end{minipage}%
        }%
        \subfigure[PP is eliminated.]{
        \begin{minipage}[t]{0.49\linewidth}
        \centering
        \includegraphics[width=0.99\textwidth]{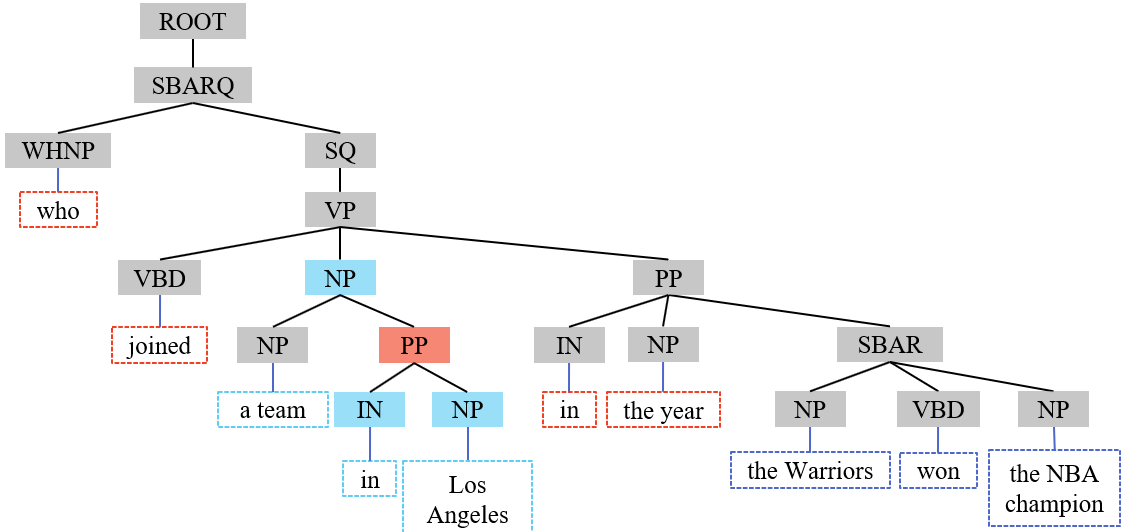}
        \end{minipage}%
        }%

    \subfigure[PP is eliminated.]{
        \begin{minipage}[t]{0.49\linewidth}
        \centering
        \includegraphics[width=0.99\textwidth]{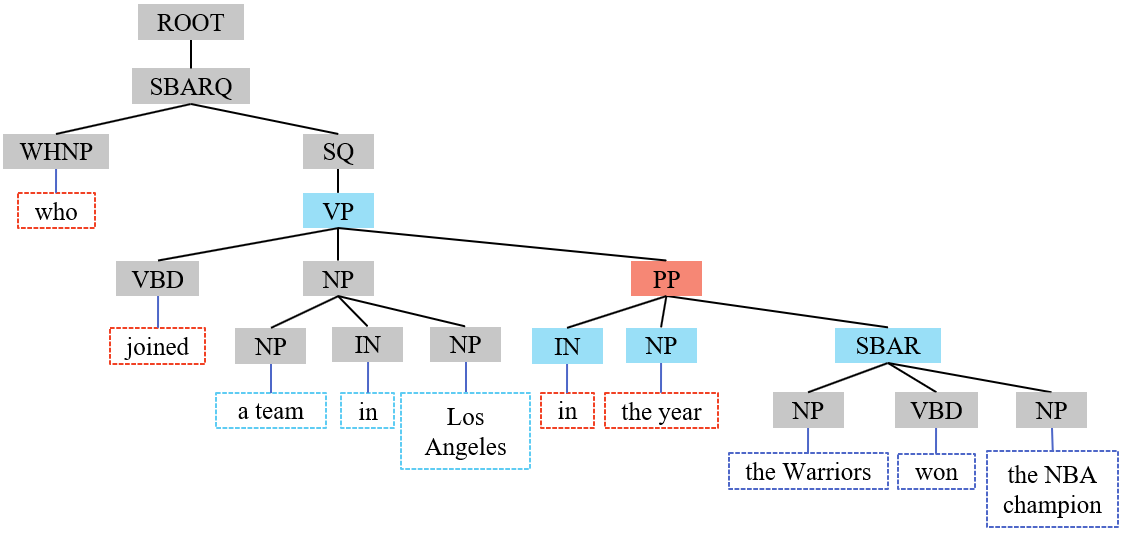}
        \end{minipage}%
        }%
        \subfigure[NP is retained.]{
        \begin{minipage}[t]{0.49\linewidth}
        \centering
        \includegraphics[width=0.99\textwidth]{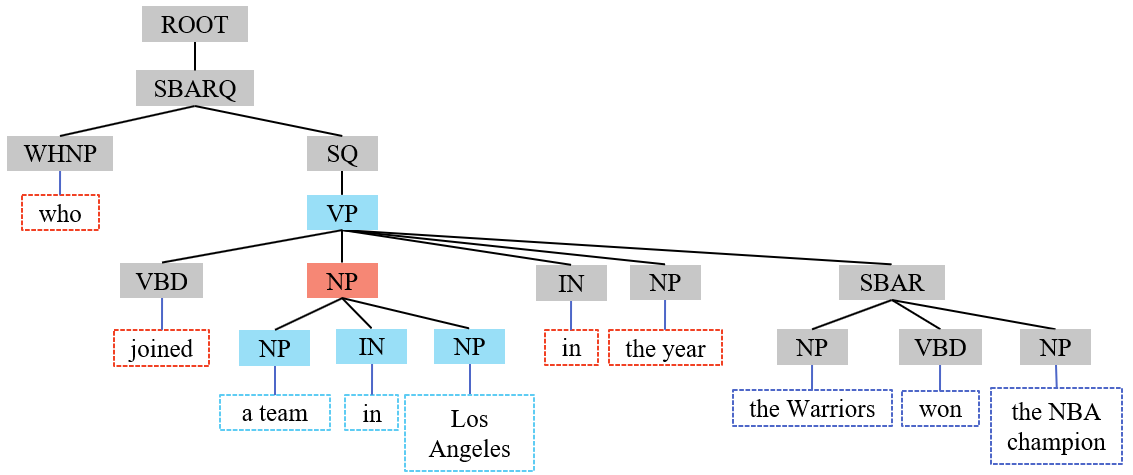}
        \end{minipage}%
        }%

    \subfigure[VP is eliminated.]{
        \begin{minipage}[t]{0.49\linewidth}
        \centering
        \includegraphics[width=0.99\textwidth]{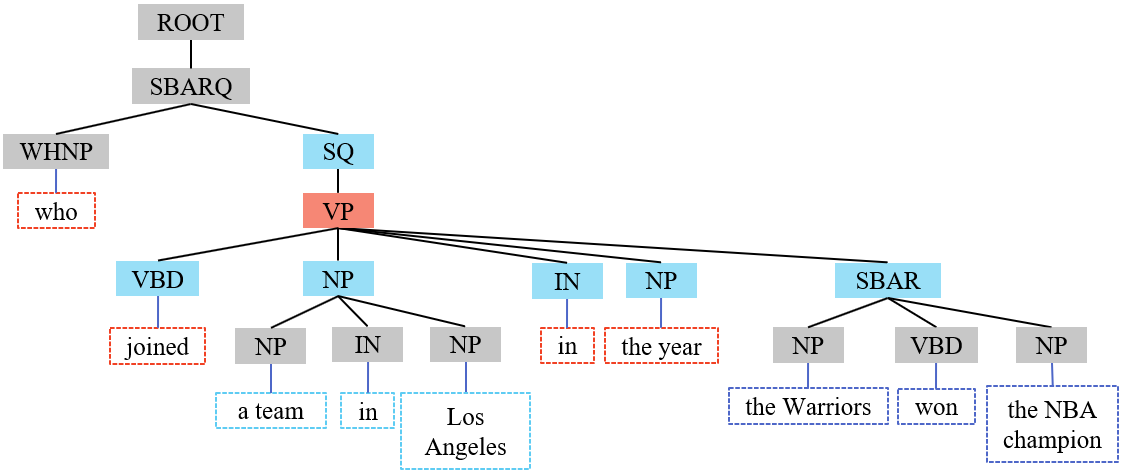}
        \end{minipage}%
        }%
        \subfigure[SQ is eliminated.]{
        \begin{minipage}[t]{0.49\linewidth}
        \centering
        \includegraphics[width=0.99\textwidth]{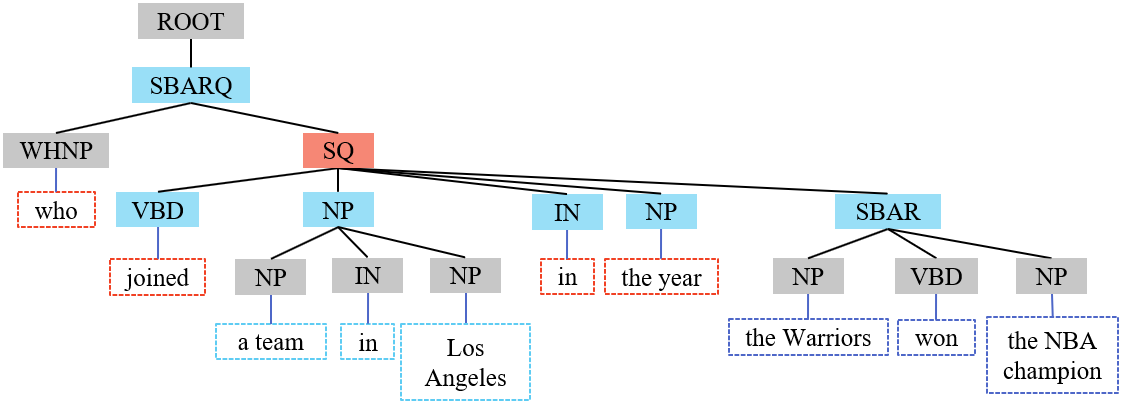}
        \end{minipage}%
        }%

        \subfigure[SBARQ is eliminated.]{
            \begin{minipage}[t]{0.49\linewidth}
            \centering
            \includegraphics[width=0.99\textwidth]{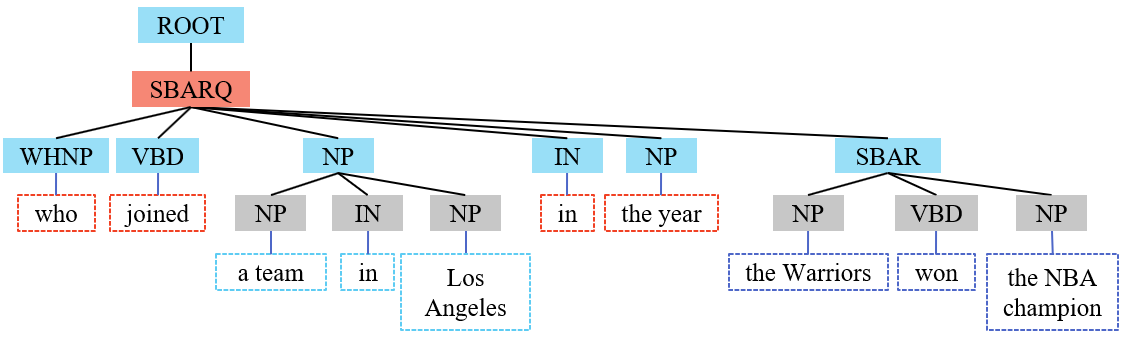}
            \end{minipage}%
            }%
            \subfigure[Elimination ends.]{
            \begin{minipage}[t]{0.49\linewidth}
            \centering
            \includegraphics[width=0.99\textwidth]{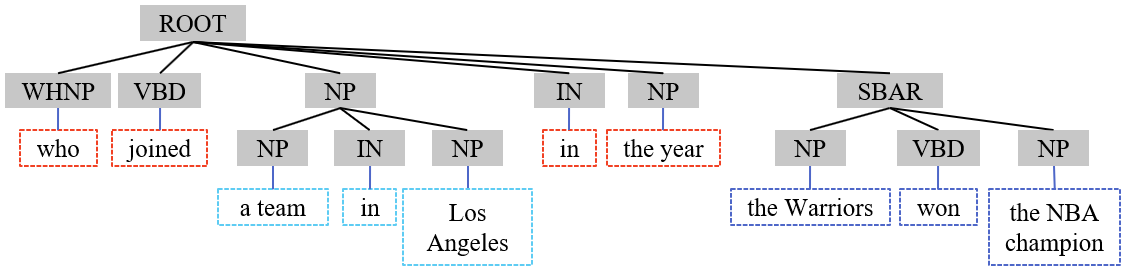}
            \end{minipage}%
            }%

    \centering
    \caption{Node elimination process.}
    \label{fig:elimination}
    \end{figure*}

Next we start eliminating nodes from the the bottom to the top as well as from the right to the left.
Note that we start the elimination operation from the third-to-last layer of the tree.
For each selected node, we extract a subtree that contains this node (colored in red) and its neighbor nodes (colored in blue). 
This subtree is fed into a classifier.
The output of the classifier determines whether to eliminate this node.
Figure~\ref{fig:elimination} shows the specific elimination process.
The previously selected node will no longer be selected, e.g., the node ``SBAR'' in Figure~\ref{fig:elimination} (f).

\begin{figure*}
    
    \centering
    \includegraphics[width=0.89\textwidth]{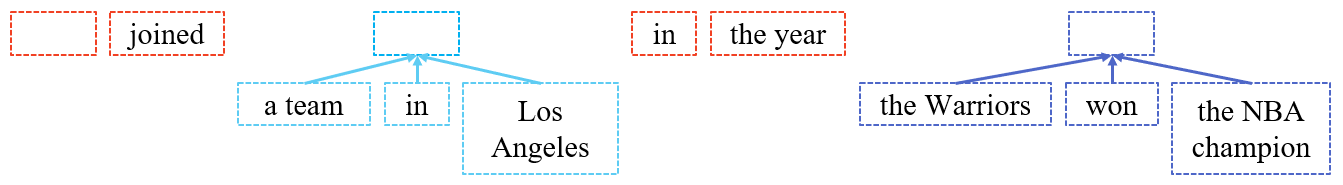}
    \caption{NL fact tree. The red, blue and purple blank nodes are placeholder nodes.}
    \label{fig:3}
    
\end{figure*}

After the elimination process, we delete non-leaf nodes and retain the hierarchical structure of leaf nodes. 
For the continuous nodes in the lower-layer, set a common placeholder node in the upper-layer.
For example, the continuous nodes \texttt{a team}, \texttt{in} and \texttt{Los Angeles} in Figure~\ref{fig:elimination} (j) will be connected to a common placeholder node~(i.e., the blue node in Figure~\ref{fig:3}).
The interrogative pronouns, e.g., \texttt{who} are also replaced directly with placeholders.
Now, the NL fact tree is constructed~(see Figure~\ref{fig:3}).
It satisfies 1) the leaf nodes are words or phrases of the question; and 2) if the leaf nodes share the same parent, they belong to the same fact.

\clearpage

\begin{table*}[]
    
    \centering
    \large
    \renewcommand{\arraystretch}{1.53}
    \resizebox{1\textwidth}{!}{
    \begin{tabular}{l|l|l}\toprule[1.5pt]
    ID & Template & Mode\\\toprule[0.5pt]
    1	&	who win $\left \{  \right \} $ award in the time $\left \{  \right \} $	&	1	\\\hline
    2	&	who is sibling of $\left \{  \right \} $	&	1	\\\hline
    3	&	what is the profession of $\left \{  \right \} $	&	1	\\\hline
    4	&	what is the country of $\left \{  \right \} $	&	1	\\\hline
    5	&	what political party did $\left \{  \right \} $ join	&	1	\\\hline
    6	&	who is the spouse of $\left \{  \right \} $	&	1	\\\hline
    7	&	what is the gener of $\left \{  \right \} $	&	1	\\\hline
    8	&	who was educated at $\left \{  \right \} $ until $\left \{  \right \} $	&	1	\\\hline
    9	&	when did $\left \{  \right \} $ die	&	1	\\\hline
    10	&	where was $\left \{  \right \} $ born in	&	1	\\\hline
    11	&	who work at the place $\left \{  \right \} $	&	1	\\\hline
    12	&	who was nominated for the prize $\left \{  \right \} $ in the time $\left \{  \right \} $	&	1	\\\hline
    13	&	who is the mother of whose father is $\left \{  \right \} $	&	2	\\\hline
    14	&	who is the father of whose mother is $\left \{  \right \} $	&	2	\\\hline
    15	&	who win award $\left \{  \right \} $ in the time when $\left \{  \right \} $ win the $\left \{  \right \} $	&	2	\\\hline
    16	&	who is the father of who has ever won $\left \{  \right \} $	&	2	\\\hline
    17	&	who is the child of who has ever won $\left \{  \right \} $	&	2	\\\hline
    18	&	who was nominated for $\left \{  \right \} $ in the time when $\left \{  \right \} $ win the $\left \{  \right \} $	&	2	\\\hline
    19	&	what political party did the father of $\left \{  \right \} $ join	&	2	\\\hline
    20	&	what is the profession of the person who has ever won the $\left \{  \right \} $	&	2	\\\hline
    21	&	who died in the place where $\left \{  \right \} $ born in	&	2	\\\hline
    22	&	who born in the place where $\left \{  \right \} $ died in	&	2	\\\hline
    23	&	which field did the person who has ever educated at $\left \{  \right \} $ work for	&	2	\\\hline
    24	&	who is the spouse of the person who born in $\left \{  \right \} $	&	2	\\\hline
    25	&	who is the father of the person who born in $\left \{  \right \} $	&	2	\\\hline
    26	&	what is the country of the person whose father is the one has ever won the prize $\left \{  \right \} $	&	3	\\\hline
    27	&	who born in the place where the father of $\left \{  \right \} $ died in	&	3	\\\hline
    28	&	who died in the place where the mother of $\left \{  \right \} $ born in	&	3	\\\hline
    29	&	who was educated at the school where the person who won the prize $\left \{  \right \} $ was also educated at	&	3	\\\hline
    30	&	who is the child of the person whose father is the one who is the sibling of $\left \{  \right \} $	&	3	\\\hline
    31	&	who work in the field that the person from the country $\left \{  \right \} $ work for	&	3	\\\hline
    32	&	who join the political party that the person from the country $\left \{  \right \} $ has erver joined	&	3	\\\hline
    33	&	when did the person from $\left \{  \right \} $ won the prize that $\left \{  \right \} $ has ever won	&	4	\\\hline
    34	&	who joined a team in $\left \{  \right \} $ in the year $\left \{  \right \} $ won the NBA championship	&	4	\\
    
    \midrule[1.5pt]  
    \end{tabular}}
    \caption{List of question templates and their fact combination mode. The curly braces $\left \{  \right \} $ indicate the entities to be filled. }
    \label{tab:questiontemplate}
    \end{table*}



\end{document}